\documentclass[10pt,twocolumn,letterpaper]{article}

\usepackage[pagenumbers]{cvpr} 

\usepackage{graphicx}
\usepackage{amsmath}
\usepackage{amssymb}
\usepackage{booktabs}
\usepackage{amssymb}
\usepackage{multirow}
\usepackage{color, colortbl}
\usepackage{algorithm}
\usepackage{algpseudocode}

\usepackage[pagebackref,breaklinks,colorlinks]{hyperref}

\usepackage[capitalize]{cleveref}
\crefname{section}{Sec.}{Secs.}
\Crefname{section}{Section}{Sections}
\Crefname{table}{Table}{Tables}
\crefname{table}{Tab.}{Tabs.}

\definecolor{Gray}{gray}{0.9}

\def\cvprPaperID{****} 
\def\confName{CVPR}
\def\confYear{2022}

\begin{document}

\title{Recurrent Glimpse-based Decoder for Detection with Transformer}

\author{Zhe Chen$^1$\hspace{1.4cm} Jing Zhang$^1$\hspace{1cm} Dacheng Tao$^{2,1}$\\
$^1$ The University of Sydney, Australia \hspace{0.5cm} 
$^2$ JD Explore Academy, China\\
{\tt\small \{zhe.chen1,jing.zhang1\}@sydney.edu.au; dacheng.tao@gmail.com}
}
\maketitle

\begin{abstract}
Although detection with Transformer (DETR) is increasingly popular, its global attention modeling requires an extremely long training period to optimize and achieve promising detection performance. Alternative to existing studies that mainly develop advanced feature or embedding designs to tackle the training issue, we point out that the Region-of-Interest (RoI) based detection refinement can easily help mitigate the difficulty of training for DETR methods. Based on this, we introduce a novel REcurrent Glimpse-based decOder (REGO) in this paper. In particular, the REGO employs a multi-stage recurrent processing structure to help the attention of DETR gradually focus on foreground objects more accurately. In each processing stage, visual features are extracted as glimpse features from RoIs with enlarged bounding box areas of detection results from the previous stage. Then, a glimpse-based decoder is introduced to provide refined detection results based on both the glimpse features and the attention modeling outputs of the previous stage. In practice, REGO can be easily embedded in representative DETR variants while maintaining their fully end-to-end training and inference pipelines. In particular, REGO helps Deformable DETR achieve 44.8 AP on the MSCOCO dataset with only 36 training epochs, compared with the first DETR and the Deformable DETR that require 500 and 50 epochs to achieve comparable performance, respectively. Experiments also show that REGO consistently boosts the performance of different DETR detectors by up to 7$\%$ relative gain at the same setting of 50 training epochs. 
Code is available via \url{https://github.com/zhechen/Deformable-DETR-REGO}.
\end{abstract}

\section{Introduction}
Object detection aims to locate and recognize foreground objects from images. In recent years, deep learning has made rapid development in object detection. With deep convolutional neural networks \cite{he2016deep, simonyan2014very, liu2021Swin,zhang2020resnest,hu2018squeeze}, various powerful detectors have been developed \cite{ren2016faster, lin2017feature, tian2019fcos,chen2022sasa}. 
In general, modern detectors produce redundant results and require Non-Maximum Suppression (NMS) to reduce the redundancy in detection.
Different from this popular paradigm, Detection with Transformer (DETR) \cite{carion2020end} applied Transformer \cite{vaswani2017attention} for detection and is the first fully end-to-end detector that avoids the need for NMS. In particular, a Transformer is a powerful attention-based encoder-decoder pipeline for translating an input sequence to the target sequence. By formulating the detection task as a direct set prediction problem, the authors of DETR managed to translate visual features into a set of detection results based on the global attention modeling of a Transformer. 
Despite benefits, the DETR suffers from a difficult training problem. Using MS COCO dataset \cite{lin2014microsoft}, the original DETR requires 500 training epochs to  obtain promising performance, while the other popular detectors like FPN \cite{lin2017feature} only require less than 36 epochs to get similar results. Even using a machine with 8 powerful V100 GPUs, a DETR detector costs more than 10 days to finish the training \cite{carion2020end}.

\begin{figure}[t]
\centering
\includegraphics[width=\linewidth]{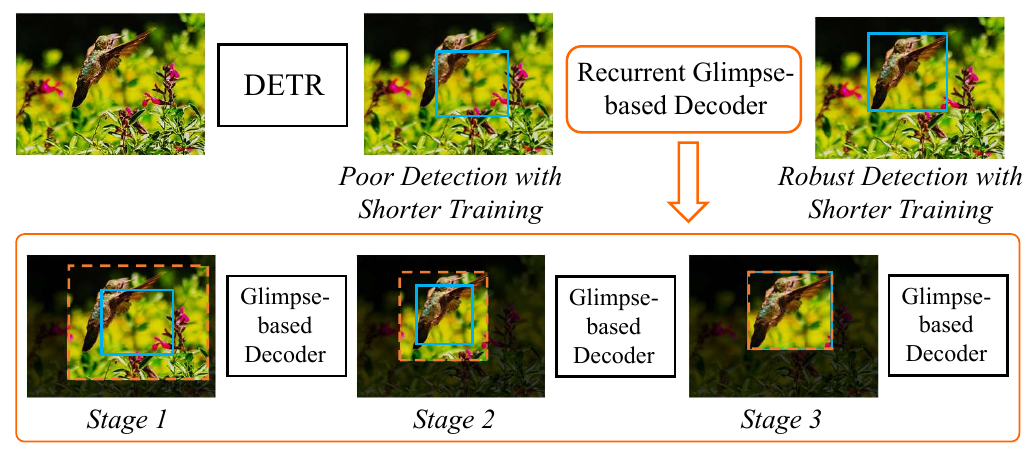}
  \caption{Concept of the proposed recurrent glimpse-based decoder (REGO) for augmenting the training of attention modeling in Detection with Transformer (DETR). Using original DETR results, the REGO performs a multi-stage Region-of-Interest (RoI) based attention modeling refinement procedure by gradually focusing on more accurate areas. In each stage, glimpse features are extracted and a glimpse-based decoder is employed to provide refined detection outputs based on both the glimpse features and the attention modeling output of the previous stage.
  The REGO maintains the fully end-to-end pipeline of different DETR methods and can improve their training performance promisingly. 
  }%
  \label{fig:title}
\end{figure}

By addressing the training problem, researchers found that the lack of effective locality modeling could affect the training of attention modeling in DETR methods. For example, Zhu \textit{et al.} \cite{zhu2020deformable} analyzed that the Transformer would distribute almost uniform attentional weights to all features initially. It is then necessary to apply long training epochs to make the Transformer learn to focus on sparse and meaningful local areas. To tackle this issue, researchers developed advanced multi-scale feature encoding \cite{dai2017deformable, gao2021fast} and object embedding designs \cite{meng2021conditional, wang2021anchor} to improve the locality modeling in Transformer before final detection, so that the attention of Transformer can be trained more efficiently and the detection results can be improved properly. 

Different from existing methods, we propose that the training of the attention modeling in DETR can be easily improved based on Region-of-Interest (RoI). More specifically, considering local areas around bounding boxes detected by DETR as RoIs that may contain objects, we can directly restrict the attention of DETR by only focusing on these RoIs. 
Therefore, modeling the features within RoIs can help introduce more locality inductive biases in DETR and thus improve its training efficiency effectively. 
In fact, researchers have demonstrated that gradual refinements according to RoIs can boost training and detection performance for two-stage \cite{girshick2015fast, ren2016faster} and multi-stage detectors \cite{ren2016faster, cai2018cascade}. Nevertheless, these multi-stage detection methods mainly follow RCNN detection methodology \cite{girshick2014rich} for training and inference which still requires NMS.
To our best knowledge, the RoI-based refinement for attention modeling in DETR has rarely been studied.

To develop a proper RoI-based DETR refinement method, we take inspiration from the glimpse mechanism as studied in the work \cite{mnih2014recurrent} which extracted features from a few selected local areas of different scales as glimpses and applied a recurrent network to encode the glimpse information. Similar to DETR, this glimpse method also formulates the visual understanding as a sequence translation task and has proven to be effective for image recognition. We follow this mechanism and propose a novel recurrent glimpse-based decoder (REGO) module to help existing DETR methods relieve training difficulty and improve detection performance. 

The proposed REGO module refines DETR with multi-stage processing. Taking detection { and attention modeling outputs} of the original DETR as initial states, each stage of REGO first extracts glimpse features from local areas surrounding the detected bounding boxes. Then, a Transformer decoder is employed to translate the glimpse features based on previous attention modeling outputs into augmented attention modeling outputs and refined detection results. For early stages, we extract glimpse features from the local areas at larger scales \textit{w.r.t.} the detected bounding box areas, enabling the incorporation of rich contexts to boost the detection that can possibly be unreliable in early stages. After multiple stages of processing, the REGO { performs a coarse-to-fine RoI-based refinement which is shown to be} effective for improving the training of different DETR methods.  

To sum up, the contributions of this paper are three-fold:
\begin{itemize}
    \item We proposed { a novel RoI-based refinement module that can effectively tackle the difficult training problem for the attention modeling in DETR and improve detection performance}.
    \item The REGO is easy-to-implement and is a complementary module that can be embedded in different DETR variants. It keeps the fully end-to-end detection pipeline of DETR while accelerating convergence and improving detection performance for different DETR methods effectively.
    \item Extensive experiments show that the REGO helps deliver promising performance using only 36 training epochs with a DETR pipeline, which is 13$\times$ shorter than the first DETR method. Moreover, REGO also consistently boosts the performance of different DETR methods by up to 7\% relative gain using the same 50 training epochs.
\end{itemize}

\section{Related Work}
\noindent\textbf{Object Detection}
Modern detectors \cite{ren2016faster,lin2017feature,he2017mask,chen2021recursive,chen2021shape} generally make dense detection of objects appeared on the images. For example, the widely used regional proposal network (RPN) \cite{ren2016faster} scans every location on the feature map of the backbone like ResNet\cite{he2016deep} and generates proposal windows that may cover foreground objects. This produces plenty of redundant proposals, \eg, an object can be covered by different but highly overlapped proposals, which is disadvantageous to make sparse predictions. To alleviate this problem, one-stage methods \cite{redmon2016you, liu2016ssd, lin2017focal, law2018cornernet} develop augmented networks to ensure that they can directly provide compact detection results. Two-stage methods \cite{ren2016faster, he2017mask} attempt to refine the proposal bounding boxes based on the features extracted with RoIPooling \cite{girshick2015fast} or RoIAlign \cite{he2017mask}. Nevertheless, both one-stage and two-stage detectors rely on the hand-designed NMS procedure to remove redundancy, which is heuristic and separated from the end-to-end learning pipeline, leading to many inaccurate predictions remained after NMS. 
Alternatively, recently introduced detection with Transformer \cite{carion2020end} can provide a set of object detection results without requiring NMS. 
However, DETR suffers from frustratingly difficult training problem. 

\noindent\textbf{Improvement of Transformer in Computer Vision}
The training difficulty of DETR is a common issue in Transformer-based computer vision methods \cite{dosovitskiy2020vit, touvron2021training,wang2021fp}. 
By addressing the training problem, many researchers found that locality modeling is important for improving the training of attention in DETR \cite{tay2020efficient, xu2021vitae, zhang2022vitaev2}. Some methods \cite{ramachandran2019stand} developed advanced local window-based method for improving efficiency. In object detection, researchers mainly develop advanced feature encoding and embedding designs to help tackle this problem. The Deformable DETR \cite{zhu2020deformable} applied deformable operations \cite{dai2017deformable} to better focus on a few local areas at different scales in the Transformer. The method SMCA \cite{gao2021fast} introduced multi-scale co-attention to improve DETR with refined local representations. In addition, other studies like Conditional DETR \cite{meng2021conditional} and Anchor DETR \cite{wang2021anchor} tend to improve the spatial embedding in Transformer to help accelerate training. These two methods enhance the locality modeling of Transformer by making attention focus on potentially valuable areas on the image learned with positional embeddings. Unlike these methods that require careful designs, we argue that the RoIs which naturally correspond to local areas can also improve the training of attention modeling in DETR.
{ A more related method is the iterative refinement used in Deformable DETR\cite{zhu2020deformable}. We note that this method does not use RoIs and it mainly improves performance by re-using all the regression outputs of DETR. Our RoI method can improve attention modeling and is orthogonal to this method. Experiments show that the cooperation of this method and our REGO achieves state-of-the-art performance. 
}

\noindent\textbf{RoI-based Improvement for Object Detection}
Researchers have proved that the detection results can be progressively improved by refining classification and localization \textit{w.r.t.} RoIs \cite{yang2016craft,cai2018cascade,gidaris2016attend,chen2018context,liu2020asts}. For example, MR CNN \cite{gidaris2015object} introduced an iterative procedure to alternate between scoring and bounding box refinement based on RoIs. 
The CascadeRCNN \cite{cai2018cascade} repeated the RoI-based detection head of the Faster RCNN\cite{ren2016faster} several times for refinement. Despite effectiveness, such type of RoI-based refinement methodology can not be directly applied to the fully end-to-end pipeline of DETR because they rely on different optimization goals and still require NMS. 
More recently, some methods, like Efficient DETR \cite{yao2021efficient}, TSP-RCNN \cite{sun2021rethinking}, and SparseRCNN\cite{peize2020sparse}, also uses RoIs to achieve improved performance with a Transformer and can also avoid the NMS. However, we argue that these methods are still based on the typical two-stage detection pipeline like Faster RCNN\cite{ren2016faster} and they only apply Transformer mainly to approximate NMS. These methods do not directly tackle the difficult training problem for attention modeling in DETR.

In summary, exploring end-to-end RoI-based refinement { for improving the training of attention modeling} in DETR remains a missing part in literature.

\begin{figure*}[t]
\centering
\includegraphics[width=0.88\linewidth, height=0.33\textheight]{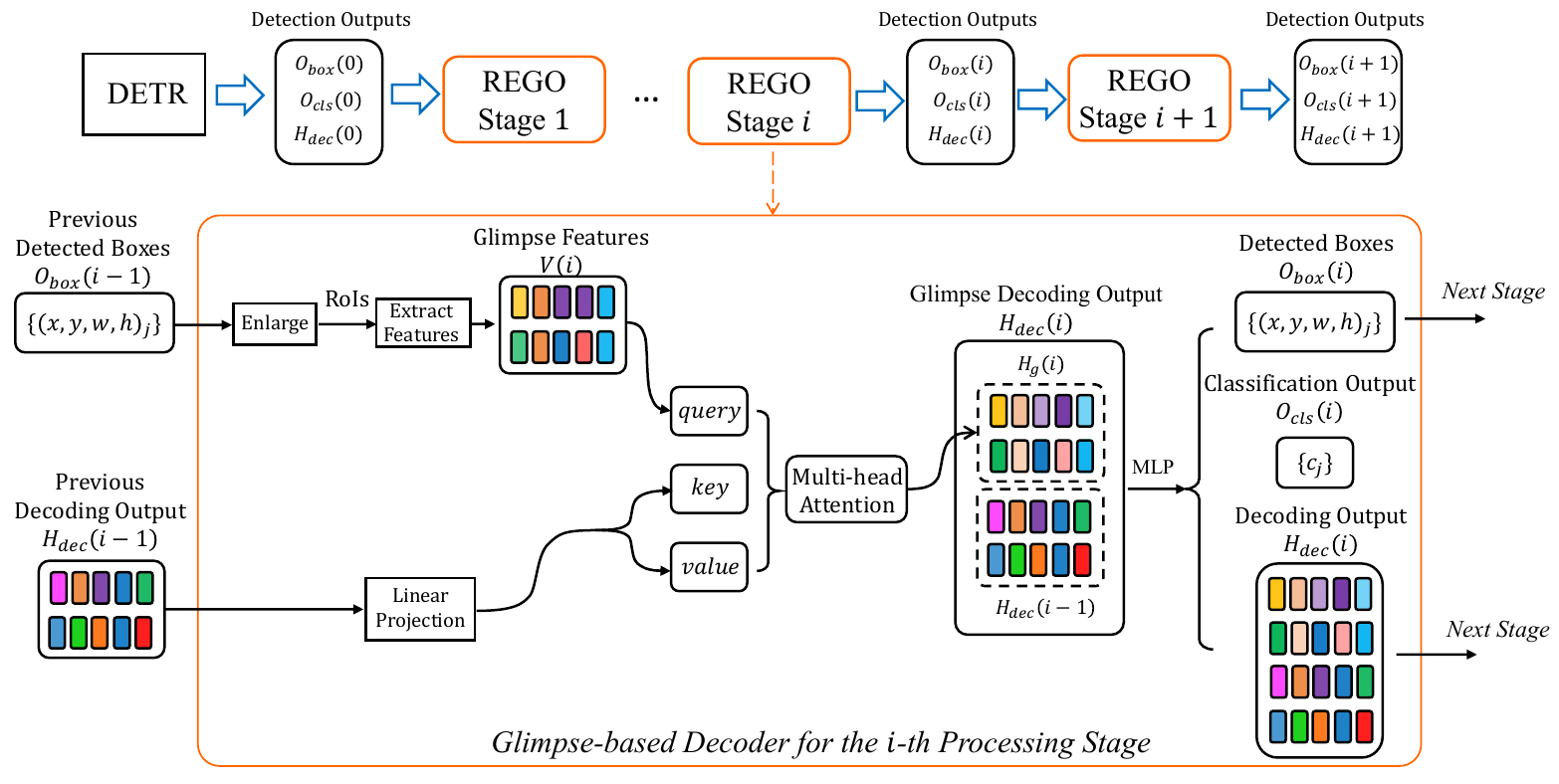}
  \vspace{-0.2cm}
  \caption{The overview of the REGO (top row) and the detailed structure of the $i$-th processing stage in REGO. 
  }%
  \label{fig:method}
  \vspace{-0.2cm}
\end{figure*}

\section{Preliminary}
Here, we briefly review the DETR. More details can be found at \cite{vaswani2017attention,carion2020end}. 

\noindent\textbf{Multi-head Attention} 
Multi-head attention deals with \textit{query}, \textit{key}, and \textit{value} inputs. It correlates query and key and then aggregates values according to the correlation results. 
Following \cite{vaswani2017attention}, the multi-head attention splits features into different 'heads' and performs self-attention or cross-attention in each head. The features of different heads will be concatenated together and fed into a linear projection to obtain the final output. Formally, to help describe a multi-head attention, we suppose that ${X}_q \in \mathbb{R}^{L_q\times C}$ is a query tensor where $L_q$ refers to its sequence length and $C$ is its feature dimension. We follow the formulations of DETR \cite{carion2020end} and unify the key and value into the same tensor: ${X}_{kv}\in \mathbb{R}^{L_{kv}\times C}$ which is the the key-value sequence of length $L_{kv}$. The multi-head attention, abbreviated as "$\mathcal{A}$", can be formulated as:
\begin{equation}
    \mathcal{A}({X}_q, {X}_{kv}) = W_\mathcal{A} \big[\mathcal{A}({X}^1_{q},{X}^1_{kv}), \ldots, \mathcal{A}({X}^M_{q}, {X}^M_{kv}) \big],
    \label{eq:ma}
\end{equation}
where $W_\mathcal{A}\in \mathbb{R}^{C\times C}$ is a trainable linear projection matrix, $M$ is the number of heads, and $[ \ldots]$ refers to concatenation operation. The ${X}^i_{q}\in \mathbb{R}^{L_q\times C'}$ and ${X}^i_{kv}\in \mathbb{R}^{L_{kv}\times C'}$ are query and key-value tensors of the $i$-th head ($i=1,\ldots M$), respectively, where $C'=\frac{C}{M}$. In each head, the following operation is performed:
\begin{equation}
     \mathcal{A}({X}^i_{q}, {X}^i_{kv}) = \mathcal{A}^i_{qkv} {X}^i_{kv},
\end{equation}
where $\mathcal{A}^i_{qkv}$ represents the attentional weights: $\mathcal{A}^i_{qkv} = Softmax(\frac{{X}^i_{q} ({X}^i_{kv})^T}{\sqrt{C'}})$.

\noindent\textbf{DETR Pipeline} 
The DETR applies an encoder-decoder pipeline to translate the input features into a set of detection results. During training, Hungarian matching \cite{kuhn1955hungarian} is performed to assign the detection results with the most matched ground-truths. The encoder-decoder consists of a visual feature encoding phase and a detection result decoding phase. 
The feature encoding investigates the relations between visual features from different locations. It applies several encoding layers to augment the encoded representation. We suppose the backbone network extracts features into: ${X}\in\mathbb{R}^{H\times W\times C}$ where $H,W$ represents the height and width, respectively, and $C$ is the feature dimension. 
In each encoding layer, a multi-head \emph{self-attention} module is employed, that is, query, key, and value tensors are the same: ${X}_q = {X}_{kv}$. The input feature ${X}$ also integrates a positional embedding to encode position information. Suppose the output of the encoding phase is ${H}_{enc} \in \mathbb{R}^{HW \times C}$. Then, detection decoding phase perform detection based on ${H}_{enc}$. It begins from object query embeddings ${E}_{box}\in\mathbb{R}^{N_d\times C}$ and applies \emph{cross-attention} as described in Eq. \eqref{eq:ma} with ${H}_{enc}$ for making predictions. The $N_d$ here represents the number of predicted objects. Suppose the decoded feature is $H_{dec}$, then $H_{dec}=\mathcal{A}({E}_{box}, {H}_{enc})$. With the ${H}_{dec}$, the decoding phase performs classification and bounding box coordinate regression, obtaining ${O}_{cls}\in \mathbb{R}^{N_d \times N_c}$ and ${O}_{box}\in \mathbb{R}^{N_d \times 4}$, respectively, where $N_c$ represents the number of object categories. To this end, we have:
\begin{equation}
    \left\{
    \begin{array}{ll}
      {O}_{cls} = F_{cls}({H}_{dec})\\
      {O}_{box} = F_{box}({H}_{dec})
    \end{array}
      \right.,
      \label{eq:orig}
\end{equation}
where $F_{cls}$ and $F_{box}$ are functions that map the decoded feature ${H}_{dec}$ into the desired outputs respectively. The two functions are implemented based on linear projection and multi-layer perception, respectively.

\section{Recurrent Glimpse-based Decoder}
Different from existing methods, we propose a recurrent glimpse-based decoders (REGO) to perform RoI-based detection refinement method for improving attention modeling in DETR. The REGO consists of two major components. The first one is a multi-stage recurrent processing structure that progressively augments attention modeling outputs and improves the detection of DETR, and the second one is the glimpse-based decoder that is used in each stage to explicitly perform the refinement. Figure \ref{fig:method} shows the detailed pipeline. 

\subsection{Multi-stage Recurrent Processing}
Built upon detection results and attention decoding outputs from original DETR, we propose a recurrent processing pipeline to help the DETR gradually attend to more meaningful areas to avoid long training periods for optimizing the attention of DETR. In general, the proposed recurrent processing structure is a multi-stage pipeline. In each stage, previously detected bounding boxes are used to obtain RoIs for extracting glimpse features. Then, glimpse features are translated according to previous attention decoding outputs into refined attention decoding outputs for describing detected objects. The refined attention decoding outputs can provide improved detection results. Thus, for the $i$-th processing stage, we propose to detect objects according to:
\begin{equation}
    \left\{
    \begin{array}{ll}
      {O}_{cls}(i) = F_{cls}({H}_{dec}(i))\\
      {O}_{box}(i) = F_{box}({H}_{dec}(i)) + {O}_{box}(i-1)
    \end{array}
      \right.,
    \label{eq:rec}
\end{equation}
where $O_{cls/box}(i)$ represent the classification and bounding box regression outputs of the $i$-th recurrent processing stage, respectively, and $H_{dec}(i)$ represents the refined attention of this stage after decoding. Then, to obtain a proper representation of $H_{dec}(i)$, we use the following formulation: 
\begin{equation}
    H_{dec}(i) = [H_g(i), H_{dec}(i-1)], 
    \label{eq:dec}
\end{equation}
where $H_g(i)$ is the translated glimpse features according to $H_{dec}(i-1)$, and $[ \ldots]$ refers to concatenation operation. Reusing $H_{dec}(i-1)$ in Eq. \eqref{eq:dec} not only improves the attention of previous stages, but also help maintain consistency in the produced detection results across different stages, which could help reduce the variations in the Hungarian matching loss in later stages
The study\cite{sun2021rethinking} has proven that reducing the randomness of the matching loss is beneficial for accelerating convergence. The calculation of translated glimpse features $H_g(i)$ will be discussed with more details in the next section. For the first stage where $i=0$, we use the outputs of original DETR, as described in Eq. \eqref{eq:orig}, to represent $O_{cls}(0)$, $O_{box}(0)$, and $H_{dec}(0)$.

\subsection{Glimpse-based Decoder}
During the $i$-th processing stage, the glimpse-based decoder collects visual features from areas around the detected bounding boxes $O_{box}(i-1)$ from the previous stage. It then performs cross-attention to model the relations between the collected features and previous attention outputs and compute translated glimpse features $H_g(i)$ of current stage. 

In particular, we denote the extracted visual features as $V(i)$ for the $i$-th stage, terming as the glimpse feature. Then, we translate it according to the previous attention outputs into a refined attention modeling outputs for detection. Multi-head cross-attention is applied to fulfill the translation, \ie,
\begin{equation}
    H_g(i) = \mathcal{A}(V(i), H_{dec}(i-1)).
    \label{eq:gdec}
\end{equation}
Note that we use the attention outputs from the last layer of the decoder in original DETR to define $H_{dec}(0)$. It is also worth mentioning that either the $V(i)$ or the $H_{dec}(i-1)$ can be used as the query in $\mathcal{A}$. Both settings can correlate glimpse features with previous attention outputs properly and can all improve the training of DETR. We simply found that above formulation achieves 0.5 point higher in AP on COCO dataset\cite{lin2014microsoft}.

To extract the glimpse features $V(i)$, we perform the following operation based on $O_{box}(i-1)$:
\begin{equation}
    V(i) = f_{ext}\Big(X, R\big(O_{box}(i-1), \alpha(i)\big) \Big),
    \label{eq:ext}
\end{equation}
where the function $f_{ext}$ represents the feature extraction operation, $R$ represents the RoI computation, and $\alpha(i)$ a scalar factor. In particular, the function $R$ computes RoIs by enlarging the areas of bounding boxes detected by $O_{box}(i-1)$ with a factor of $\alpha$. Then, we use the RoIAlign\cite{he2017mask} technique to implement $f_{ext}$. 
The symbol $X$ here represents the features obtained with the backbone network. 

Since the original detection results could be unreliable at first, we tend to extract glimpse features from a larger area around each detection result for refinement in early stages, so that contexts can be incorporated and target objects can be properly captured within the glimpse areas. In later processing stages, we gradually narrow the area for extracting glimpse features to achieve more precise detection with more local details. In other words, the $\alpha(i)$ in REGO starts from a large number and then decreases its value for later stages of the REGO. The detailed setting of $\alpha(i)$ can be found in the following section. 

\subsection{Implementation Details}
\label{subsec:implementationdetails}
The proposed REGO is a plug-and-play module for different DETR methods. 
It only has two major hyperparameters, \textit{i.e.} the number of recurrent stages and the enlarging ratios $\alpha$ of each stage. To reduce the manual tuning efforts, we unify the two hyperparameters into a single one. More specifically, we constrain that the last recurrent stage has an enlarging ratio equals to 1. Then, when we add a new recurrent stage before the last stage, we increase the enlarging ratio by 1 for the added stage. In other words, if we use 3 recurrent stage, then $\alpha(3), \alpha(2), \alpha(1)=3,2,1$, respectively. Therefore, we only need to investigate the influence of number of recurrent stages.
In addition, we follow the original DETR and apply auxiliary losses to enhance the training of intermediate outputs of the glimpse-based decoders and apply LayerNorm \cite{ba2016layer} to help regularize the decoded glimpse representation $H_{dec}(i)$. 

For each recurrent stage of the REGO, we use the decoder architecture of the original DETR for glimpse feature translation, but we do not use encoders and only use 2 decoding layers for a decoder. In decoders, the self-attention of encoders brings marginal benefits but consumes more computational resources, \textit{e.g.} for REGO-DeformabelDETR-R50, adding the self-attention layers for all stages only improves AP, AP$_{50}$, and AP$_{75}$ by 0.1, -0.1, 0.2, respectively, while introducing around 4 more GFLOPs and 9M more parameters. 
Without encoders, the complexity of the decoder in REGO is much smaller than the decoder used in the original DETR methods. The complexity analysis is presented in the experiment section. 
Besides the number of stages, we present other implementation details as follows.  Firstly, we follow the default settings of the
RoIAlign\cite{he2017mask} and uses a 7 by 7 window for feature extraction. In addition, when extracting glimpse features, we attempt to use features from different levels of backbone in both multi-scale and single-scale DETR methods, but note that we do not use the FPN \cite{lin2017feature} to save costs. Also, the number of RoIs depends on the output of DETR and the number of stages. We will present more details in supplementary materials.

\begin{table*}[t!]
\centering
\resizebox{0.98\linewidth}{!}{
\begin{tabular}{l| c |c| c|cc  c c c | c | c }
\hline
Detectors & Backbone & Epochs & AP & AP$_{50}$ & AP$_{75}$ & AP$_{S}$ & AP$_{M}$ & AP$_{L}$ & GFLOPs & \#Params (M) \\ %
\hline
\hline
FCOS\cite{tian2019fcos} & R50&36&41.0& 59.8& 44.1 &26.2& 44.6& 52.2 & 177 & 32\\
Faster RCNN - FPN \cite{lin2017feature} & R50&36&40.2 &61.0 &43.8 &24.2& 43.5& 52.0 & 180 & 42\\
Faster RCNN - FPN \cite{lin2017feature}& R101& 36& 42.0& 62.5 &45.9 &25.2 &45.6 &54.6 & 246 & 61\\
Mask RCNN \cite{he2017mask}& X101& 36 & 44.5 & 64.9 & 48.7 & 27.6 & 48.3 & 57.7&457 & 102 \\
Cascade Mask RCNN\cite{cai2018cascade, chen2019hybrid} & X101& 36 & 46.6 & 65.1 & 50.6 & 29.3& 50.5 & 60.1 & 627 & 135\\
TSP-RCNN \cite{sun2021rethinking} & R50 & 96 & 45.0 &64.5 &49.6 &29.7 &47.7 &58.0 & 188 & -\\
Efficient DETR \cite{yao2021efficient} & R50 & 36 & 44.2& 62.2& 48.0 &28.4 &47.5& 56.6 & 159 & 35\\
Sparse RCNN\cite{peize2020sparse} & R50 & 36 &44.5 &63.4 &48.2 &26.9& 47.2& 59.5 & - & -\\
\hline
DETR \cite{carion2020end} & R50 & 500 & 42.0 & 62.4 & 44.2 & 20.5 & 45.8 & 61.1 & 86 & 41 \\ %
DETR-DC5 \cite{carion2020end} & R50 & 500 &  43.3  & 63.1 &45.9  & 22.5 & 47.3 & 61.1& 187 & 41\\
UP-DETR\cite{dai2021up} &R50 & 300 & 42.8 &63.0 &45.3& 20.8& 47.1 &61.7& 86 & 41\\
Conditional DETR\cite{meng2021conditional} & R50 & 50 & 40.9  &61.8  &43.3  &20.8  &44.6  &59.2 & 90& 44\\
Anchor DETR \cite{wang2021anchor} & R50 & 50 &44.2& 64.7 &47.5 &24.7 &48.2& 60.6& 151 & -\\
SMCA \cite{gao2021fast} & R50 & 50 & 43.7 &63.6& 47.2 &24.2&47.0 &60.4& 152&40\\
SMCA \cite{gao2021fast} & R101 & 50 & 44.4& 65.2& 48.0& 24.3 &48.5 &61.0 & 218&58\\
\hline
DETR$^*$ \cite{carion2020end,zhu2020deformable,meng2021conditional} $^\dag$ &  & 50 & 39.3 & 60.3 & 41.4 & 18.5 & 42.4 & 57.5 & 88 & \multirow{2}{*}{44}\\
DETR$^*$-DC5 \cite{carion2020end,zhu2020deformable, meng2021conditional} $^\dag$ & \multirow{-2}{*}{R50} & 50 &  41.3 & 62.8 & 43.6 & 21.0 & 44.5 & 59.4 & 189 & \\
\rowcolor{Gray}
REGO-DETR$^*$ \textbf{(ours)} &  & 50 &  42.3 & 60.5 & 46.2 & 26.2 & 44.8 & 57.5 & 112 & \\
\rowcolor{Gray}
REGO-DETR$^*$-DC5 \textbf{(ours)} & \multirow{-2}{*}{R50} & 50 & 
44.0 & 62.6 &  47.8 & 26.5 & 45.2& 62.9  & 213 & \multirow{-2}{*}{58}\\ 
\multirow{2}{*}{Deformable DETR\cite{zhu2020deformable}} & \multirow{2}{*}{R50}& 36$^\dag$ &42.7& 61.4 &46.7 &25.9 &46.2 &56.6 & \multirow{2}{*}{173} & \multirow{2}{*}{40}\\ 
&  & 50 &43.8 &62.6 &47.7 &26.4 &47.1 &58.0 & &\\
\rowcolor{Gray} 
& & 36 & 44.8 & 63.8 & 48.7 & 27.0 & 48.0 & 60.2&  & \\ 
\rowcolor{Gray}
\multirow{-2}{*}{REGO-Deformable DETR \textbf{(ours)}}& \multirow{-2}{*}{R50} & 50 & 45.9 & 65.2 & 49.7 & 27.6 & 48.9 & 61.5 & \multirow{-2}{*}{190} &\multirow{-2}{*}{54} \\  
 & R50 & 50  & 46.4 & 65.3 & 50.6 & 30.0 & 49.8 & 61.4& 173 & 40\\ %
 & R101  & 50& 47.2 & 66.6 & 51.1 & 28.5 & 50.9 & 62.4 & 240 &59\\ %
\multirow{-3}{*}{Deformable DETR$^{**}$ \cite{zhu2020deformable}$^\dag$} & X101 &50 & 47.7 & 67.2 & 51.4 & 29.3 & 51.2 & 62.8 &417&105\\ %
\rowcolor{Gray}
 & R50 & {50} & 47.6 & 66.8 &51.6&29.6&50.6&62.3& 190& 54 \\  
\rowcolor{Gray}
 &R101 & 50 & 48.5 & 67.0 & 52.4 & 29.5 & 52.0 & 64.4 & 257 &73 \\
\rowcolor{Gray}
\multirow{-3}{*}{REGO-Deformable DETR $^{**}$ \textbf{(ours)} } & X101  & 50
 & \textbf{49.1} & \textbf{67.5} &\textbf{53.1} & \textbf{30.0} &\textbf{52.6}&\textbf{65.0} & 434&119\\ 
\hline
\end{tabular}
}
\caption{Results of different detectors on the MS COCO \textit{val} split. Baseline results are shaded. $^*$ Improve with 300 queries, reference points, and focal loss \cite{zhu2020deformable,meng2021conditional}. $^{**}$ Improve with iterative box refinement and two-stage processing. $^\dag$ Reproduced using released code. }
\label{tab:exp-all}
\vspace{-0.2cm}
\end{table*}

\section{Experiment}
\subsection{Setup}
We follow existing DETR methods \cite{carion2020end} and perform evaluation using the MS COCO \cite{lin2014microsoft} dataset which has 118k training images and 5k validation images. We follow the MS COCO protocol and report the performance using the evaluation metrics of average precision (AP), AP at 0.5, AP at 0.75, and AP for small, medium, and large objects. The validation set is mainly used for evaluation. 

We apply our method on the original DETR \cite{carion2020end} and Deformable DETR \cite{zhu2020deformable} using their released codes. For training, we follow the original settings of the released codes for fair comparison, except that we also perform experiments with much fewer training epochs. For example, the original DETR detectors adopt 500 or 50 training epochs, while we mainly evaluate our method with 50 or 36 training epochs. 

\subsection{Performance Evaluation}
In this section, we perform comprehensive comparison between the current DETR methods and our method. Table \ref{tab:exp-all} shows the overall results 
on MS COCO \textit{val} dataset. In particular, we thoroughly investigate the performance of applying REGO on different DETR methods using different backbone networks and different training epochs.

\noindent\textbf{Comparison with different DETR methods}
We have applied our proposed REGO on two major DETR detectors for evaluation. This include the vanilla DETR \cite{carion2020end} method improved with 300 queries, reference points, and focal loss as described by \cite{zhu2020deformable} and the Deformable DETR \cite{zhu2020deformable}. We also presented the reported performance of RCNN-based methods \cite{tian2019fcos, ren2016faster, lin2017feature, cai2018cascade, peize2020sparse, sun2021rethinking} and other DETR variants \cite{dai2021up, wang2021anchor, gao2021fast, meng2021conditional, yao2021efficient}.
From the results in Table \ref{tab:exp-all}, we can observe that our method consistently improves different R50-based baseline methods by around 2 points in AP using 50 epochs. For example, using the original DETR, we boosts the performance from 39.3 AP to 42.3 AP at 50 training epochs. 
Moreover, by further cooperating with iterative box refinement and two-stage processing when using the Deformable DETR as baseline, the REGO helps improve the AP from 46.4 to 47.6 which is the highest score in all the compared DETR methods trained with 50 epochs and R50 backbone. This demonstrates that our proposed REGO is effective for improving DETR by gradually attending to more accurate object areas with RoIs.

\noindent\textbf{Comparison at Fewer Training Epochs}
By applying the REGO on the Deformable DETR, we also compare the detection performance obtained at 36 training epochs used by many traditional detection methods \cite{lin2017feature,tian2019fcos}. According to the Table \ref{tab:exp-all}, we help the Deformable DETR achieve 44.8 AP using 36 training epochs while the original Deformable DETR only achieves 42.7 with 36 epochs. Under the same training period, our method also helps surpass FPN and FCOS greatly. We also performed an extra experiment of REGO-DeformableDETR-X101 also trained with 36 epochs, obtaining 48.1, 67.4, 52.0 in AP, AP$_{50}$, and AP$_{75}$, respectively, which are higher than the CascadeRCNN\cite{cai2018cascade}, proving that our REGO can reduce training costs for DETR effectively. 

\noindent\textbf{Comparison with Different Backbones}
We also investigate the effectiveness of REGO on different backbone networks, including R50\cite{he2016deep}, R101\cite{he2016deep}, and X101\cite{xie2017aggregated}. Besides the improvements over R50 network, the results in Table~\ref{tab:exp-all} also show that REGO continues to improve the baseline DETR method with both R101 and X101 networks promisingly. In particular, with X101 backbone network, our Deformable DETR + REGO detector achieves the highest AP among many state-of-the-art object detectors.

\begin{figure}[t]
\centering
\includegraphics[width=0.8\linewidth]{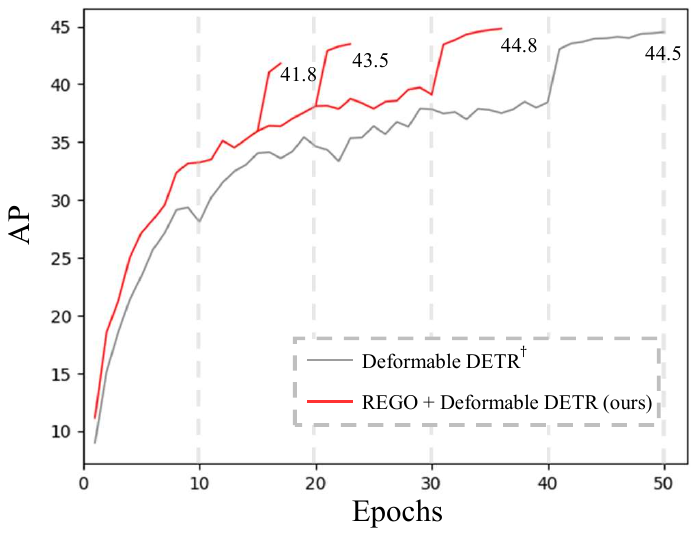}
  \caption{Convergence curves of Deformable DETR on the \textit{val} set for whether using the proposed REGO. For REGO, we explore convergence performance by reducing the learning rate at the 15-th, 20-th, and 30-th epoch. $\dag$: Reproduced using released code.}%
  \label{fig:curve}
\end{figure}

\noindent\textbf{Convergence Analysis}
We further study the impact of REGO on actual convergence. Fig. \ref{fig:curve} shows the detailed convergence curves of the Deformable DETR and the Deformable DETR with REGO. It shows that the REGO effectively speeds up the convergence and boosts the model performance promisingly comparing to the baseline. In particular, REGO helps achieve comparable performance with the baseline using only 30 epochs, \ie, 40\% less than the complete  50 training epochs used in the baseline. Comparing to the first DETR which requires 500 epochs, the REGO can help reduce about 94\% of the total training period. 

\begin{figure}[t]
\centering
\includegraphics[width=0.8\linewidth]{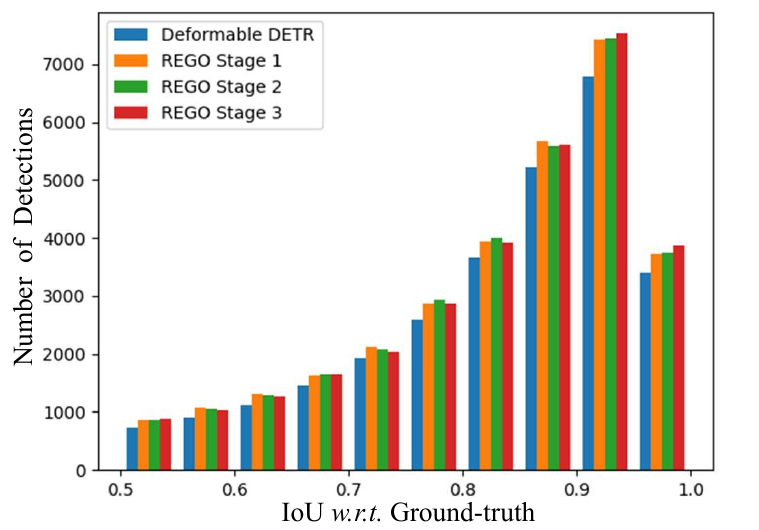}
  \caption{Histogram of correct detection results on the \textit{val} set at different settings, \ie, different IoU \textit{w.r.t.} ground-truths and in different REGO stages. Note that the number of correct detection results of different stages share a similar amount ($\sim$ 30k boxes). }%
  \label{fig:iou}
  \vspace{-0.2cm}
\end{figure} 

\noindent\textbf{Complexity Analysis}
The extra computational complexity brought by the REGO is around 17 GFLOPs, which is only around 10\% of the complexity of a Deformable DETR-R50 model while bringing around 28\% acceleration in training (36 epochs \textit{v.s.} 50 epochs). Furthermore, the complexity of our method remains the same when using larger and deeper backbone networks like R101 and X101 because of the identical implementation. The extra complexity \textit{w.r.t.} these larger backbone network-based DETR are only around 7\%, while the REGO brings more improvement rather than increasing depth of backbone networks. For example, the X101 improves R50 with 1.3 points in AP (46.4 to 47.7) for Deformable DETR at the cost of another 244 GFLOPs, while the REGO achieves similar results (47.6) with only extra 17 GFLOPs using the R50 backbone. 
Furthermore, we can also show in the supplementary materials that the Deformable DETR trained with REGO can already achieve around 1 point higher AP even without using REGO during inference, which means that the REGO directly helps original DETR learn better attention and offers detection improvement during inference \emph{for free}.

\begin{table}[t!]
\centering
\resizebox{\linewidth}{!}{
\begin{tabular}{l | c|cc  c c c }
\hline
Glimpse Stages  & AP & AP$_{50}$ & AP$_{75}$ & AP$_{S}$ & AP$_{M}$ & AP$_{L}$ \\ 
\hline
Deformable DETR \cite{zhu2020deformable} & 43.8 &62.6 &47.7 &26.4 &47.1 &58.0\\ 
\hline
1 Stage ($\alpha$ = 1) &45.1  &63.0 & 46.4 & 24.7 & 46.0 & 60.0\\
2 Stage ($\alpha$ = 2, 1) &45.6 & 65.1 & 49.3 & 27.4 & 48.7 & 61.2\\ %
3 Stage ($\alpha$ = 3, 2, 1)  &45.9 & 65.2 & 49.7 & 27.6 & 48.9 & 61.5 \\ 
4 Stage ($\alpha$ = 4, 3, 2, 1) & 45.9 & 65.5 & 50.3 & 28.5 & 49.0 & 61.1\\ %
\hline
\end{tabular}
}
\caption{ Hyper-parameter study of the number of stages in REGO. Deformable DETR \cite{zhu2020deformable} is used as baseline.}
\label{tab:exp-stage}
\end{table}

\begin{table}[t!]
\centering
\resizebox{0.9\linewidth}{!}{
\begin{tabular}{l | c|cc  c c c }
\hline
Glimpse Scale  & AP & AP$_{50}$ & AP$_{75}$ & AP$_{S}$ & AP$_{M}$ & AP$_{L}$ \\ 
\hline
1x & 45.9 & 65.2 & 49.7 & 27.6 & 48.9 & 61.5\\ 
\hline
1.5x  & 45.8 & 65.0 & 50.1 & 27.6 &48.9 & 61.3  \\ %
\hline
2x & 45.7 & 65.0 & 49.9 & 27.8 & 48.6 & 59.9 \\ %
\hline
\end{tabular}
}
\caption{ Hyper-parameter study of the glimpse scale in REGO. REGO is implemented with 3 stages.}
\label{tab:exp-scale}
  \vspace{-0.2cm}
\end{table}

\begin{figure*}[t!]
\centering
\includegraphics[width=\linewidth,height=0.33\textheight]{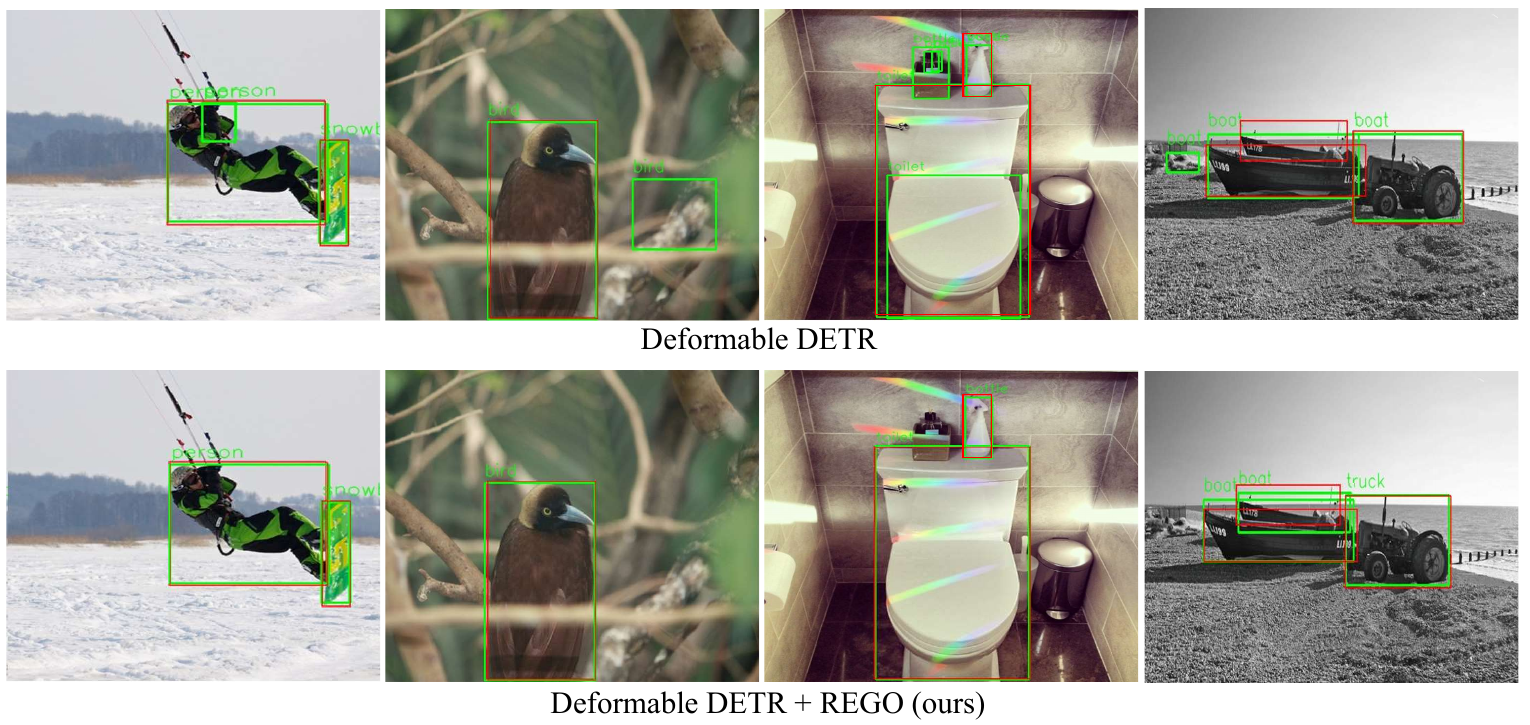}
  \vspace{-0.2cm}
  \caption{Visual detection results of the baseline Deformable DETR \cite{zhu2020deformable} and its variant with REGO. Green boxes are detection results while red boxes are ground-truths.}%
  \label{fig:vis}
  \vspace{-0.2cm}
\end{figure*} 

\subsection{Ablation Study}
\noindent\textbf{Analysis of Different REGO Stages}
We first present the detection performance of applying different numbers of REGO stages in Table \ref{tab:exp-stage}. The evaluated stages range from 1 to 4 where the glimpse scale in each stage varies accordingly as described in Section~\ref{subsec:implementationdetails}. We also present the baseline Deformable DETR result as reported in the paper \cite{zhu2020deformable}. 
The results show that the REGO with different numbers of stages improves the baseline performance greatly. A single processing stage can increase the mAP by more than 1 point. Applying more stages leads to further improvement. The performance of 3 and 4 stages is the best among compared settings. Although REGO with 4 stages achieves favorable performance, its improvement over the setting of 3 stages is marginal, implying that adding more than 3 processing stages in REGO could result in diminished benefits. We also disentangle the enlarging ratio from the 3-stage setting, \textit{i.e.} making $\alpha$=1,1,1, and only obtain 45.3 in mAP, showing that the glimpse design is useful.

We also study the quality of detection results from different stages in a 3-stage REGO module. A histogram chart of the numbers of correct detection results at different settings of Intersect-over-Union (IoU) \textit{w.r.t.} ground-truths is shown in Fig. \ref{fig:iou}. A detection result is correct only if its IoU to a ground-truth is higher than 0.5 and its predicted label aligns with the ground-truth. In addition, the numbers of total detection results for different stages are similar, \ie, around 30k boxes. Therefore, with the similar amount of correct detection results, the chart shows that all the REGO stages (red bars) help produce more accurate detection results than the baseline (blue bars) and their counterparts with fewer stages (yellow and green bars) for the right-most two groups of detection results. For example, the results of REGO with 3 stages contain more correct detection results whose IoU scores \textit{w.r.t.} ground-truths are higher than 0.9. These results demonstrates that REGO with more stages continues to refine the detection by focusing on objects in the coarse-to-fine ROIs and learning better feature representations.

\noindent\textbf{Analysis of Different Scales of Glimpse Area}
Table \ref{tab:exp-scale} shows the performance comparison of using different scales of glimpse area. The 1x, 1.5x, and 2x in the table represent the ratios of enlarging glimpse scales. For example, if using 2x and the default glimpse scales are 3.0, 2.0, 1.0 times larger than the previously detected bounding boxes, the actual glimpse scales are 6.0, 3.0, 2.0 times larger, respectively. We can find that the 1x setting already achieves the highest AP, and other settings achieve comparable but slightly lower AP. This suggests it is inappropriate to enlarge glimpse areas aggressively for implementing REGO. 

\subsection{Qualitative Results}
We present some visual detection results to better illustrate the impact of REGO. We choose Deformable DETR\cite{zhu2020deformable} with R50 as baseline. Fig. \ref{fig:vis} shows the results. Note that we choose the detection results whose confidence scores are higher than 0.5 for better visualization. From the figure, we can observe that the REGO indeed helps reduce both false positive and false negative results for the baseline method. Besides, the REGO can also help investigate the relations between different detected bounding boxes with the help of glimpse-based decoders. We will present some visual examples of the object relations learned with REGO in the supplementary materials.

\section{Conclusion}
We introduce a novel and effective technique, called REcurrent Glimpse-based decOder (REGO), to improve the Detection with Transformer (DETR) methods. By incorporating recurrent processing structure and learning glimpse features from coarse-to-fine RoIs, the REGO shows to both accelerate the convergence speed and boost the detection performance of different DETR methods consistently. We hope this study can contribute to future research on end-to-end and efficient detection methodologies. 

\noindent\textbf{Social Impacts and Limitations} Our method can benefit various applications like self-driving. 
A potential limitation is that we still need several GPU days for training which is environmental costly. This can be mitigated by further improving the efficiency of both our REGO and the DETR. 

\noindent\textbf{Acknowledgement.} Dr. Zhe Chen is supported by IH-180100002, and Dr. Jing Zhang is supported by ARC FL-170100117.

\appendix
\section{Appendix}
\subsection{More Descriptions}
\textbf{Processing Algorithm} In general, the REGO follows the algorithm described in Alg. \ref{alg} to process the visual features within each stage. We will release the code shortly.

\begin{algorithm}
\caption{Processing of $i$-th REGO Stage}\label{alg}
\begin{algorithmic}
\Require $H_{dec}(i-1)$, $O_{box}(i-1)$
\Ensure $H_{dec}(i)$, $O_{box}(i)$, and $O_{cls}(i)$
\State 1. Calculate RoIs by enlarging bounding box areas of $O_{box}(i-1)$ according to a scale $\alpha(i)$
\State 2. Extract glimpse features $V(i)$ based on enlarged RoIs;
\State 3. Perform multi-head attention on glimpse features $V(i)$ and $H_{dec}(i-1)$ to obtain decoded features $H_g(i)$;
\State 4. Concatenate $H_g(i)$ and $H_{dec}(i-1)$ to obtain refined attention modeling outputs $H_{dec}(i)$ of current stage;
\State 5. Predict object bounding boxes $O_{box}(i)$ and corresponding labels $O_{cls}(i)$ using $H_{dec}(i)$;
\end{algorithmic}
\end{algorithm}

\textbf{More Implementation Details} In addition to the details discussed in the paper, we would also like to mention the following aspects of implementation. In particular, regarding the extraction of multi-scale features, this can be easy for applying REGO on DETR methods like Deformable DETR \cite{zhu2020deformable} that already extract multi-scale features for attention modeling. When applying the REGO on DETR methods like the original DETR \cite{carion2020end} that only extract the single-scale feature from the last convolutional stage of the backbone, we attach 1$\times$1 convolutions on the output of different convolutional stages (stage level 2 to stage level 5) to obtain multi-scale features. In this case, the attached 1$\times$1 convolutions reduce the channel numbers to 256. Note that we do not use FPN for extracting multi-scale features to save costs. Besides, for the 'DC5' DETR variants which use single-scale features but enlarge the scale of the last convolutional stage, we still attach 1$\times$1 convolutions to extract features and reduce channel numbers, except that the features of the last convolutional stage is down-sampled to its normal scale (\textit{i.e.}, 1/32 of the input image) to save costs. 

In addition, in the REGO, we use different weight parameters to initialize the decoders of different stages, thus the decoder at each stage can be trained to be more sensitive to the glimpse features of the corresponding stage. At each stage, to stabilize training, we follow \cite{teed2020raft} and do not back propagate gradients into the outputs of the previous stage.

\begin{table}[t!]
\centering
\resizebox{\linewidth}{!}{
\begin{tabular}{l | c|cc  c c c }
\hline
Inference Method & AP & AP$_{50}$ & AP$_{75}$ & AP$_{S}$ & AP$_{M}$ & AP$_{L}$ \\ 
\hline
Deformable DETR \cite{zhu2020deformable} & 43.8 &62.6 &47.7 &26.4 &47.1 &58.0\\ 
\hline
Inference w/ REGO &45.9 & 65.2 & 49.7 & 27.6 & 48.9 & 61.5\\
Inference w/o REGO &44.9 & 65.0 & 48.7 & 26.5 & 48.4 & 61.1\\ %
\hline
\end{tabular}
}
\caption{Performance comparison of whether using REGO for inference. Deformable DETR \cite{zhu2020deformable} is used as baseline. The models related to the performance of inference with or without REGO all use REGO for training. }
\label{tab:free}
\end{table}

\begin{figure*}[t]
\centering
\includegraphics[width=\linewidth]{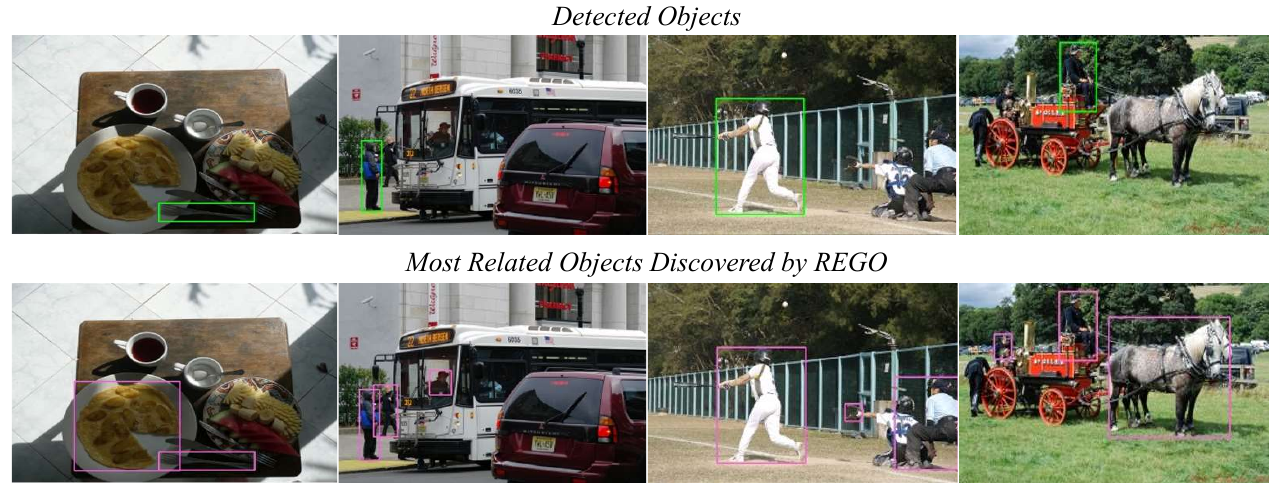}
  \caption{Illustration of using the REGO to reveal object relations. Each figure of the top row shows one of the detected objects of the previous stage, and each figure of the bottom row shows the most related detection results of the previous stage discovered by the decoder of the REGO. Results are obtained from the last stage of the REGO. 
  }%
  \label{fig:rel}
\end{figure*}

\subsection{Improve DETR For Free}
As mentioned in our paper, the proposed REGO method can improve the DETR for free during inference. This means that, after training with REGO, the obtained DETR model can still achieve improved performance {   by removing the REGO from} the detection pipeline during inference.
Table \ref{tab:free} shows the results evaluated on MS COCO \textit{val} set. We use the Deformable DETR \cite{zhu2020deformable} as our baseline DETR method.

From these results, we can find that the complete REGO method (training and inference) improves around 2 points in AP, and the Deformable DETR trained with REGO but tested without REGO still achieves around 1 point gain in AP comparing to the baseline method. This demonstrates that the REGO is effective to {   enhance} the feature of the original DETR method based on the RoI-based refinement procedure, which also suggest that the proposed REGO method can indeed improve the attention modeling in DETR during optimization. By removing the REGO, the obtained 1 point improvement does not introduce any extra complexity for inference comparing to the baseline DETR method.

\subsection{Learned Object Relations}
As described in the paper, the decoder of the REGO in each stage correlates the glimpse features extracted based on previously detected bounding boxes with the previous attention modeling outputs corresponding to the same set of detected bounding boxes.  Therefore, the obtained correlation results can reveal the relations between any two detected bounding boxes of the previous stage, and the detection results with higher correlation weights can be considered as more important for refining the attention modeling outputs and detection results.

In Fig. \ref{fig:rel}, we show some examples of the objects with the highest correlation weights \textit{w.r.t.} a detected object from the previous stage. From the presented figures, we can find that the most related objects discovered by the REGO generally have strong semantic connections. For example, in the first column, the 'fork' and 'pizza' are most correlated for refining the detection related to the 'fork' object; in the fourth column, the 'person' and the 'horses' are most correlated for refining the 'person' object who is driving the carriage. Both examples are intuitive to human as well. This illustrates that the REGO can help DETR explore the information from semantically meaningful areas without wasting attention on obviously irrelevant areas, which is beneficial for improving the training efficiency and effectiveness of attention modeling in DETR.

{\small
\bibliographystyle{ieee_fullname}
\bibliography{egbib}
}

\end{document}